# MULTI-POSE FACE RECOGNITION USING HYBRID FACE FEATURES DESCRIPTOR

I Gede Pasek Suta WIJAYA[1,2], Keiichi UCHIMURA[2] and Gou KOUTAKI[2]

**Abstract—** This paper presents a multi-pose face recognition approach using hybrid face features descriptors (HFFD). The HFFD is a face descriptor containing of rich discriminant information that is created by fusing some frequency-based features extracted using both wavelet and DCT analysis of several different poses of 2D face images. The main aim of this method is to represent the multi-pose face images using a dominant frequency component with still having reasonable achievement compared to the recent multi-pose face recognition methods. The HFFD based face recognition tends to achieve better performance than that of the recent 2D-based face recognition method. In addition, the HFFD-based face recognition also is sufficiently to handle large face variability due to face pose variations.

## I. INTRODUCTION

The face recognition has become hot topic since 20 years ago, even though reasonable results have been achieved. It happens because of wide and potential application for face recognition and each application has each highlighting and problems. For multi-pose face recognition is potentially to be implemented for surveillance in crowded environmental such as in airport or border cross. However, the multi-pose method still lack of recognition due to large face variability of single face. In addition, the face of a single person also has large variability due to wearing accessories, makeup, and hairstyle. In addition, the face recognition also has difficulty due to uncontrolled lighting on capturing the face images.

Many researchers have proposed several methods including one, two, and three-dimensional based algorithms, features based methods, appearance-based methods, as well as sub-space based methods.

This paper propose a face recognition approach using hybrid face features descriptors (HFFD). The HFFD is a face descriptor containing of rich discriminant information that is created by fusing some frequency-based features extracted

Wijaya, IGP Suta is with the Electrical Engineering department, Engineering Faculty, Mataram University, Jl. Majapahit 62 Mataram West Nusa Tenggara INDONESIA Telp. +62-370-636126; (e-mail: gpsutawijaya@te.ftunram.ac.id).
Uchimura, Keiichi is with Graduate School of Science and Technology, Kumamoto University, Kumamoto Shi, Kurokami 2-39-1, JAPAN, 860-8555, Telp. 096-342-3638 (e-mail: uchimura@cs.kumamoto-u.ac.jp).
Koutaki, Gou is with Graduate School of Science and Technology, Kumamoto University, Kumamoto Shi, Kurokami 2-39-1, JAPAN, 860-8555, Telp. 096-342-3638 (e-mail: koutaki@cs.kumamoto-u.ac.jp).

using both wavelet and DCT analysis of several different poses of 2D face images. The aims of this method is to represent the face image using dominant frequency component with still having reasonable achievement compare to the recent 2D and 3D face recognition methods, and to avoid using of 3D scanner which is an expensive tool to capture the multi-pose variations.

## II. PREVIOUS WORKS

The related works to our approach are the face recognition based on holistic or global matching methods. Zhao and Chellappa [1] have presented the comprehensive state of art of face recognition involving one and dimensional face recognition algorithms, and face recognition using sub-space methods like Principle Component Analysis (PCA) and Linear Discriminant Analysis (LDA) as well. Gao et al. [2] also have done the evaluations of large-scale Chinese face database, which consisting of 99594 images of 1040 subjects that have large diversity of the variations, including pose, expression, accessory, lighting, time, backgrounds, distance of subjects to camera, and the combined variations, using several well known methods: PCA, PCA+LDA, Gabor-PCA+LDA, and Local Gabor Binary Pattern Histogram Sequence (LGBPHS). The LGBPHS-based methods outperform over other three algorithms.

The recent techniques for 3D face recognition for handling multipose problem has been reviewed in Ref. [4] and the PCA-based algorithm for 3D face recognition has been developed in Refs. [5,6]. As presented by Zhang et al. [4], some statistical approaches such as Cook's Gaussian mixture-based Iterative Close Point algorithm, and Lee's Extended Gaussian Image model have been proposed for solving multipose problem. In addition, some algorithms based on geometric features, curvature feature-driven, profile-based face description, and a hierarchical system using range and texture information for feature extraction have been available. In addition, 3D face recognition based multi-features and multi-features fusion of face images for multipose problem were reported in Refs. [3,4]. In this case, the multi-features are extracted from 3D face image using three approaches namely: maximal curvature image, average edge image, and range image, respectively. The fusion features are built using weight linear combination. However, all of the proposed methods require 3D scanner/camera which is an expensive tool for data registration.

In addition, suppose we have large variability face images due to pose variations, as shown in Fig, 1. Our recent research[6], which did the face recognition using predictive mean linear discriminant analysis (PDLDA) with holistic features as dimensional reduction of the face image input, shows good enough performance in both off-line and real-time test for the face pose variations labeled by P$_4$, P$_5$, and P$_6$ of Fig 1. In off line test, the proposed method gave 99.52% recognition rate and in real-time test, it provided the recognition rate, false rejection rate, and false acceptance rate by about: >98%, 2%, and 4%, respectively with short processing time. Those achievements were achieved with considering the chrominance components of face image, which were extracted using YCBCr color space transformation. Wijaya et al.[7] also developed a multi-pose face recognition using fusion of scale invariant features transforms. However, it requires large memory space and longer training and querying time.

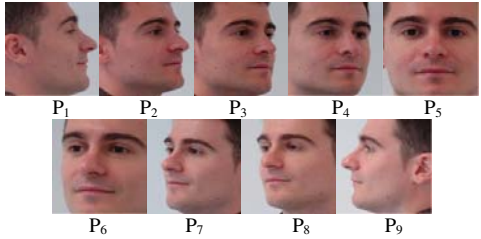

Fig. 1    Large variability of face due to pose variations.

### III.    The proposed Face Recognition

The face recognition at least consists of three main processes: features extraction, training and matching process. In the features extraction, both of training set and querying set are extracted into HFFD and save them into the file or database. The training process is required to find the projection matrix for making the HFFD more separable after the projection. Finally, the matching process is done to determine the similarity between the training and querying data.

### 2. 1 HFFD Extraction

The HFFD is a fusion of the frequency-based features of several difference face pose images which represents the multi-pose images features. The HFFD extraction is done by obtaining the frequency-based features of some face pose images and then fusing those together into one feature.

In this case, the frequency-based features are extracted by wavelet and DCT analysis. The block diagram of the frequency-based features extraction is shown in Fig. 2. The function of wavelet analysis is to extract the image into some sub-band image which each sub-band contains difference dominant frequency. From each sub-band image, the dominant frequency dominant frequency content is extracted using DCT analysis. Here, the DCT is employed because it has been proved that could extract holistic features of face image containing much useful discriminant information [6]. From the extracted DCT coefficients of each sub-band face image, small parts of them are combined into one vector called as frequency-based features.

Next, we fuse the obtained frequency-based features into features using the following procedure.

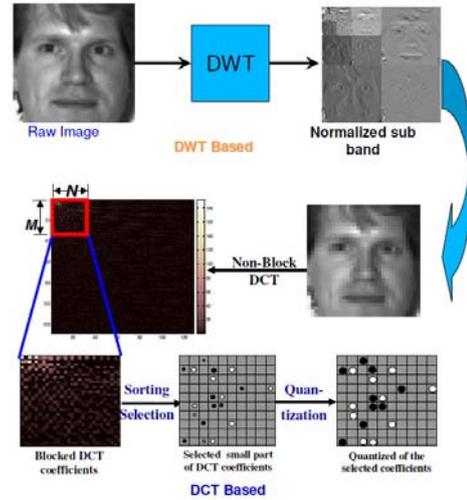

Fig. 2    Diagram block of frequency-based features extraction.

***Step 1***: Suppose select P$_4$, P$_5$, and P$_6$ of the face images in Fig. 1 as faces input set and then choose the frontal face as the basis. The basis is the principle component on fusing process. The frontal face image is chosen as basis because it has much discriminant features.

***Step 2***: Extracting the frequency-based features using algorithm in Fig. 2, which is denoted by $F_1$, $F_b$, and $F_2$ corresponding to P$_4$, P$_5$, and P$_6$, respectively.

***Step 3***: Removing the redundant features of step-by-step using intersection ($\cap$) and subtraction operation:

$$F_i^{'} = F_i - (F_i \cap F_b) \qquad (1)$$

***Step 4***: Fusing all of the non-redundant features ($F'_i$, $i$=1, 2, 3, …, $n$-1, where $n$ is number of training faces using the union ($\cup$) operation:

$$D_k = F_b \cup F_1^{'} \cup F_2^{'} \cup \cdots \cup F_{n-1}^{'} \qquad (2)$$

The extracted HFFD containing much useful discriminant information of any pose variations of face can be applied to handle the multi-pose problem.

### 2. 2 Training Process

As mentioned early, the aim this process is to find the projection matrix which makes the input features be more separable after the projection. In this case, the LDA is employed for finding the projection matrix.

The LDA works to find the optimum projection matrix ($W$) of the input data (i.e. HFFD) which is based on Fisher criterion represented by Eq. (3)[4,5].

$$J_{LDA}(W) = \arg \max_{W} \frac{\left| W^T S_b W \right|}{\left| W^T S_w W \right|} \qquad (3)$$

Where $S_w$ and $S_b$ are within class scatter and between class scatter of the training set defined as follows:

$$S_b = \sum_{k=1}^{L} P(D_k)(\mu_k - \mu_a)(\mu_k - \mu_a)^T , \qquad (4)$$

$$S_w = \frac{1}{N} \sum_{k=1}^{L} \sum_{i=1}^{N_k} (d_i^k - \mu_k)(d_i^k - \mu_k)^T , \qquad (5)$$

where $D_k$ is the HFFD of $k$-th class, $N_k$ is total samples of $k$-th class, $\mu_k$ is mean of $D_k$, $L$ is number of class, and $P(D_k)$= $N_k$ /$L$.

The optimum $W$ is determined by eigen analysis and select several large eigen vectors ($m$) which correspond to the largest eigen values. By using the obtained $W$, the HFFD of training set are projected into new space using the equation below.

$$D_k^p = W^T D_k \qquad (6)$$

where $D_k^p$ is $k$-th projected HFFD. Finally, the optimum $W$ and $D_k^p$ are saved in the database which are used in the recognition process.

### 2. 3 Matching Process

In this process, the training and querying HFFD are matched to determine the similarity score. Two techniques called as direct matching methods and integration with LDA-based matching.

The first technique classifies directly the probe HFFD with the registered/trained HFFD using the number of matching sub-band points. The matching point is determined by nearest neighbor approach. To find out the number matching point of two descriptors ($D_1 \in \Re^{r \times m}$ and $D_2 \in \Re^{r \times n}$). The classification criterion is defined according on number of matching points. The face descriptor which has the largest number of matching points is concludes as the best likeness.

In the second technique, the matching is done by calculating the distance between the training and querying projected HFFD by LDA. Suppose $D_q$ and $D_p$ are both query HFFD which are projected into LDA space using Eq. (6), then the similarity score is determined by computing Euclidean distance of $D_q$ and $D_p$s.

## IV. Experiments and Results

In order to know the performance of the proposed method, several experiments using data from challenges databases: the ITS-Lab.[5], ORL[6], and GTV databases[5] were done in PC with specification: 1.7 GHz Core-Duo Processor and 2 GB RAM.

The ITS-Lab database consists of 48 people and each person has 10 poses orientation as shown on Fig. 3(a). Total face images are 480 samples. The face images were taken by Konica Minolta camera series VIVID 900 under varying lighting condition. The ORL database was taken at different times, under varying lighting conditions with different face expressions (open/closed eyes, smiling/not smiling) and facial details (glasses/no glasses). All of the images were taken against a dark homogeneous background. The faces of the subjects are in an upright, frontal position (with tolerance for some side movement). The ORL database is a grayscale face database that consists of 40 people, mainly male. Total face images are 400 samples. The example of face pose variations of ORL database is shown in Fig. 3(b). The GTAV database is color face database, which consists of 65 people, and each person consists of 9 different face expressions. Therefore, there are 585 face images in the database. The example of face pose variations of GTAV database is shown in Fig. 3(c).

All of the experiments were performed in mentioned databases using the following assumptions: the face image size was 128 x128 pixels (28 pixels/cm); three ~ five images from each class/subject members were selected as training faces; and the training face was not included in the testing.

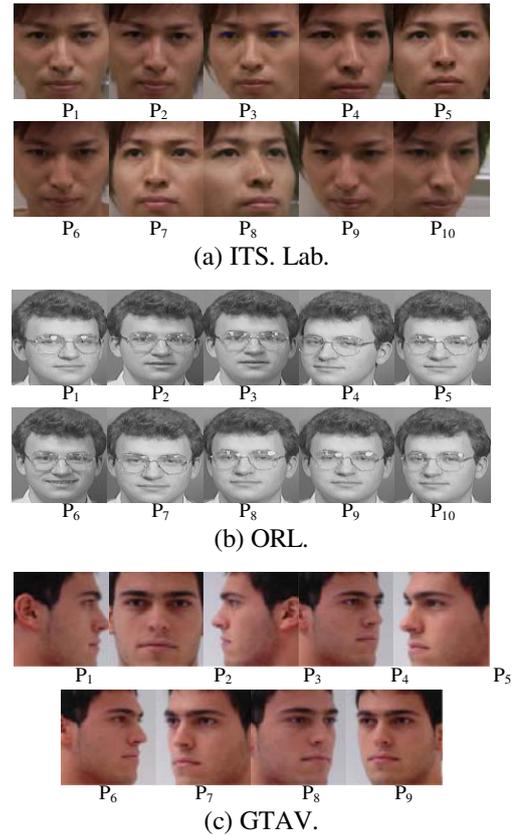

(a) ITS. Lab.

(b) ORL.

(c) GTAV.

Fig. 3    Example of face pose varition of single persons.

The first experiment was carried out to investigate the effect of HFFD on the recognition rate. In this test, we select first three images of each databases for creating HFFD and the remaining face images are selected for testing set. The experimental results show the integration of HFFD and LDA provides better recognition rate than that of single classifier (just HFFD or LDA) for all tested databases, as shown in Fig. 4. It means the HFFD is proved that contains much discriminant information of face image. In addition, the HFFD also tend to give much better performance on database containing much face variability due to face pose variations such as GTV databases[5]. It means the HFFD is proved which can handle the pose difficulty on face recognition.

In order to know the effect of number of face images for training, the second experiment was performed using 5 images per person as training set. The results shows (see Fig 5), the achievements are inline with those of the first experiments, which reprove that the HFFD provides much discriminant features of face image. In addition, the Fig 4 and Fig 5 also show that the more face images per class are considered for training set the higher recognition rate is given. In other words, the more face images per class are considered on creating the HFFD, the more rich face descriptor will be. The HFFD also reprove that it can handle the pose difficulty on face recognition which is shown by much higher improvement on large face pose variability database (such as GTV database) than that of the ITS-Lab. and ORL databases.

Regarding to the time consumption, the HFFD requires much less training time than that of LDA, as shown in Fig. 6. It caused by two factors: the HFFD size ($n$) is much less than the raw image size (less than 256 features of 128 x 128 pixels) and the eigen analysis does not become the bottle-neck of process even though the computational complexity of eigen analysis is $O(n^3)$. In addition, the HFFD time complexity do not much affect to the training time process because the fast algorithm wavelet and DCT was implemented for HFFD extractions.

The next experiment was performed to evaluate deeply the stability of the proposed method to face pose variations. The experiments were done in large and non large pose variations data: the ITS-Lab. and GTAV databases, respectively. From the ITS-Lab and GTAV databases, first three face images ($P_1 \sim P_3$ of Fig. 3(a)) and first five face images ($P_1 \sim P_5$ of Fig. 3(c)) per class were selected for creating HFFD and remaining for testing. The HFFD+LDA method provides better and more stable recognition rate than just HFFD and LDA methods for tested databases, as shown in Fig. 7. It can be achieved because the LDA projection can remove non discriminant information between the classes which make the projected HFFD by LDA become more separable than that of without LDA projection. These achievements also confirm the first and the second experimental results which the HFFD+LDA can overcame the pose variability problem. From both databases, the HFFD and HFFD+LDA methods give significant recognition rate improvement for GTAV face database containing large face variability due to pose variations. It also means the HFFD created form 2D dimensional images using frequency analysis is sufficiently for representing the large variability face images due to pose variations.

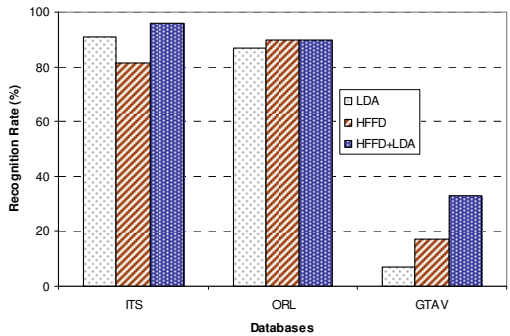

Fig. 4 Recognition rate of the proposed methods on several databases when 3 face images per class as training data.

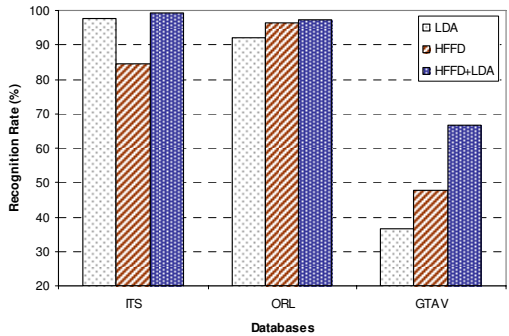

Fig. 5 Recognition rate of the proposed methods on several databases when 5 face images per class as training data.

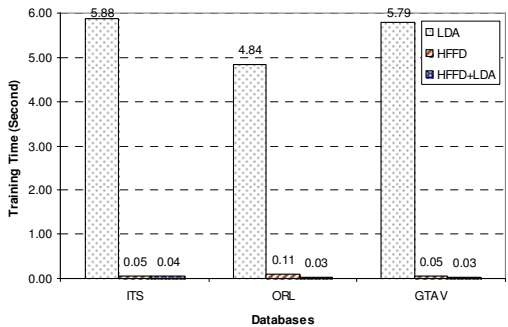

Fig. 6 Training Time of the HFFD based face recognition compared to LDA methods.

The last experiment was performed to know the performance of the proposed method compare to the recent methods such as Multi-Feature Fusion (MFF) and FSIF-based multi-pose face recognition. The MFF-based face recognition is a fusion of 3D face images combined PCA or PCALDA features cluster for multi pose face recognition[4,5]. While FSIF is the fusion of scale invariant features transform descriptor for representing the multi-pose face variations [6]. In this test, we compare three MF+PCALDA and FSIF+LDA which are best variants of those methods are used as the comparators. The experiments were carried out in the ITS-Lab face database version 1 which is 3D face database containing 40 classes which each classes consist of 10 face pose variations. The face images were captured by 3D-camera. The testing parameters were setup the same as done by Zhang et al. [4,5]: five images of each classes was chosen for training set and the remaining images as testing.

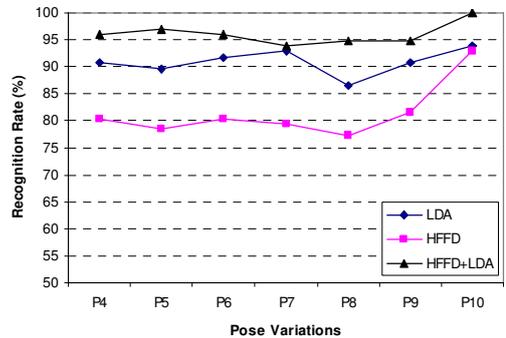

(a) ITS-Lab. database

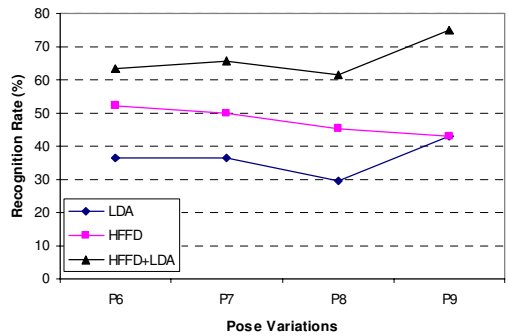

(b) GTAV database

Fig. 7 Recognition stability over the face pose variations of the HFFD based face recognition.

The HFFD+LDA provide almost the same recognition rate as FSIF+LDA and MFF+PCALDA (by 99.85%, 99.99%, and 99.98%, respectively) when training set is 5 face images per class. The HFFD+LDA and FSIF+LDA also have the same achievement for testing using 3 face images per class as shown in Table I. These results reconfirm the previous experimental results that the proposed method is an alternative approach for handling large pose variability problem. The same as FSIF+LDA methods, the proposed method does not require 3D camera sensors for making the features at all, while the MFF+PCALDA do. Event though the HFFD+LDA and FSIF+LDA have the same achievement in term of recognition rate but out proposed method spend much less training time than that of FSIF+LDA as shown in Table 1. In other words, our proposed method not only provides stable recognition rate against the poses variations but also works very fast.

In term of memory space requirement for saving the descriptor, the HFFD requires much less by about 2KB per class than the FSIF and MFF which require 100 KB and 200 KB per class respectively.

TABLE I
THE RECOGNITION RATE AND TIME CONSUMPTION OF THE PROPOSED METHODS VS. THAT OF MFF+ PCALDA AND FSIF+LDA.

| No | Methods | Recognition Rate | Time (Second) | |
|---|---|---|---|---|
| | | | Training | Querying |
| 1 | MF+PCA | 94.08 | NA | NA |
| 2 | MF+PCALDA | 99.34 | NA | NA |
| 3 | MFF+PCALDA | 99.98 | NA | NA |
| 4 | FSIF+LDA | 99.99 | 3.45 | 0.55 |
| 5 | HFFD+LDA | **99.75** | **0.04** | **0.02** |

V. CONCLUSION AND FUTURE WORKS

The HFFD is a fusion of dominant frequency features concepts which contains much useful discriminant information of face images which can be implemented for handling multi pose face recognition problem. The integration of HFFD and LDA classifier for face recognition has been proved to give sufficient and robust recognition rate against the pose variations. In addition, the HFFD+LDA method has almost the same recognition rate as that of that of recent methods but requiring much short training and querying time. In addition the HFFD also have small dimensional size which requires much less memory space for the saving and processing.

The proposed methods need to test in large size database in order to know the further robustness to handle the multi-poses problem. It also can be implemented for real time face recognition on crowded environmental.

ACKNOWLEDGMENT

I would like to send my special and great thank to Innovation Project of Kumamoto University for many supports to this research, and also my great thank to owner of ORL and GTAV databases.

**I Gede Pasek Suta Wijaya** received the B.Eng. degrees in Electrical Engineering from Gadjah Mada University in 1997, M.Eng. degrees in Computer Informatics System from Gadjah Mada university in 2001, and D.Eng. degrees in Computer Science especially on System Information from Graduate School of Science and Technology, Kumamoto University, Japan in 2010. His research interests are pattern recognition, artificial intelligence, and image processing application on computer vision.

**Keiichi Uchimura** received the B.Eng. and M.Eng. degrees from Kumamoto University, Kumamoto, Japan, in 1975 and 1977, respectively, and the Ph.D. degree from Tohoku University, Miyagi, Japan, in 1987. He is currently a Professor with the Graduate School of Science and Technology, Kumamoto University. He is engaged in research on intelligent transportation systems, and computer vision. From 1992 to 1993, he was a Visiting Researcher at McMaster University, Hamilton, ON, Canada. His research interests are computer vision and optimization problems in the Intelligence Transport System.

**Gou Koutaki** received the B.Eng., M.Eng., and Ph.D.degree from Kumamoto University, Kumamoto, Japan, in 2002, 2004, and 2007, respectively. From 2007 to 2010, he was with Production Engineering Research Laboratory, Hitachi Ltd. He is currently an Assistant Professor with the Graduate School of Science and Technology, Kumamoto University. He is engaged in research on image processing and pattern recognition of the Intelligence Transport System.